\documentclass[letterpaper, 10 pt, conference]{ieeeconf}  

\IEEEoverridecommandlockouts                              

\usepackage{graphics} 
\usepackage{amsmath} 
\usepackage{graphicx}
\usepackage{multirow}
\usepackage{makecell}
\usepackage{xcolor}
\usepackage{balance}

\usepackage{hyperref}
\hypersetup{
    colorlinks=true,
    linkcolor=black,
    filecolor=magenta,      
    urlcolor=black,
    pdftitle={Overleaf Example},
    pdfpagemode=FullScreen,
    }

\title{\LARGE \bf
RaceVLA: VLA-based Racing Drone Navigation with Human-like Behaviour
}

\author{Valerii Serpiva$^{*}$, Artem Lykov$^{*}$, Artyom Myshlyaev, Muhammad Haris Khan, Ali Alridha Abdulkarim, \\ Oleg Sautenkov and Dzmitry Tsetserukou
\thanks{$^{*}$ These authors contributed equally to this work.}
\thanks{The authors are with the Intelligent Space Robotics Laboratory, Skolkovo Institute of Science and Technology Moscow, Bolshoy Boulevard 30, bld. 1, 121205, Moscow, Russia. 
\tt \{Valerii.Serpiva, Artem.Lykov, Artyom.Myshlyaev, Haris.Khan, Ali.Abdulkarim, Oleg.Sautenkov, D.Tsetserukou\}@skoltech.ru}
}

\begin{document}

\maketitle
\thispagestyle{empty}
\pagestyle{empty}

\begin{abstract}

RaceVLA presents an innovative approach for autonomous racing drone navigation by leveraging Visual-Language-Action (VLA) to emulate human-like behavior. This research explores the integration of advanced algorithms that enable drones to adapt their navigation strategies based on the real-time environmental feedback, mimicking the decision-making processes of human pilots. The model, fine-tuned on a collected racing drone dataset, demonstrates strong generalization despite the complexity of the drone racing environments. RaceVLA outperforms OpenVLA in motion (75.0 vs. 60.0) and semantic generalization (45.5 vs. 36.3), benefiting from the dynamic camera and simplified motion tasks. However, visual (79.6 vs. 87.0) and physical (50.0 vs. 76.7) generalization were slightly reduced due to the challenges of the maneuvering in dynamic environments with varying object sizes. RaceVLA also outperforms RT-2 across all axes—visual (79.6 vs. 52.0), motion (75.0 vs. 55.0), physical (50.0 vs. 26.7), and semantic (45.5 vs. 38.8), demonstrating its robustness for real-time adjustments in complex environments. Experiments revealed an average velocity of 1.04 m/s, with a maximum speed of 2.02 m/s, and consistent maneuverability, demonstrating RaceVLA's ability to handle high-speed scenarios effectively. These findings highlight the huge potential of the RaceVLA for high-performance navigation in competitive racing contexts. The RaceVLA codebase, pretrained weights, and dataset are open-source and readily available on \textbf{https://racevla.github.io/}

\end{abstract}

\section{Introduction}


\begin{figure}[t]
    \centering
    \includegraphics[width=\linewidth]{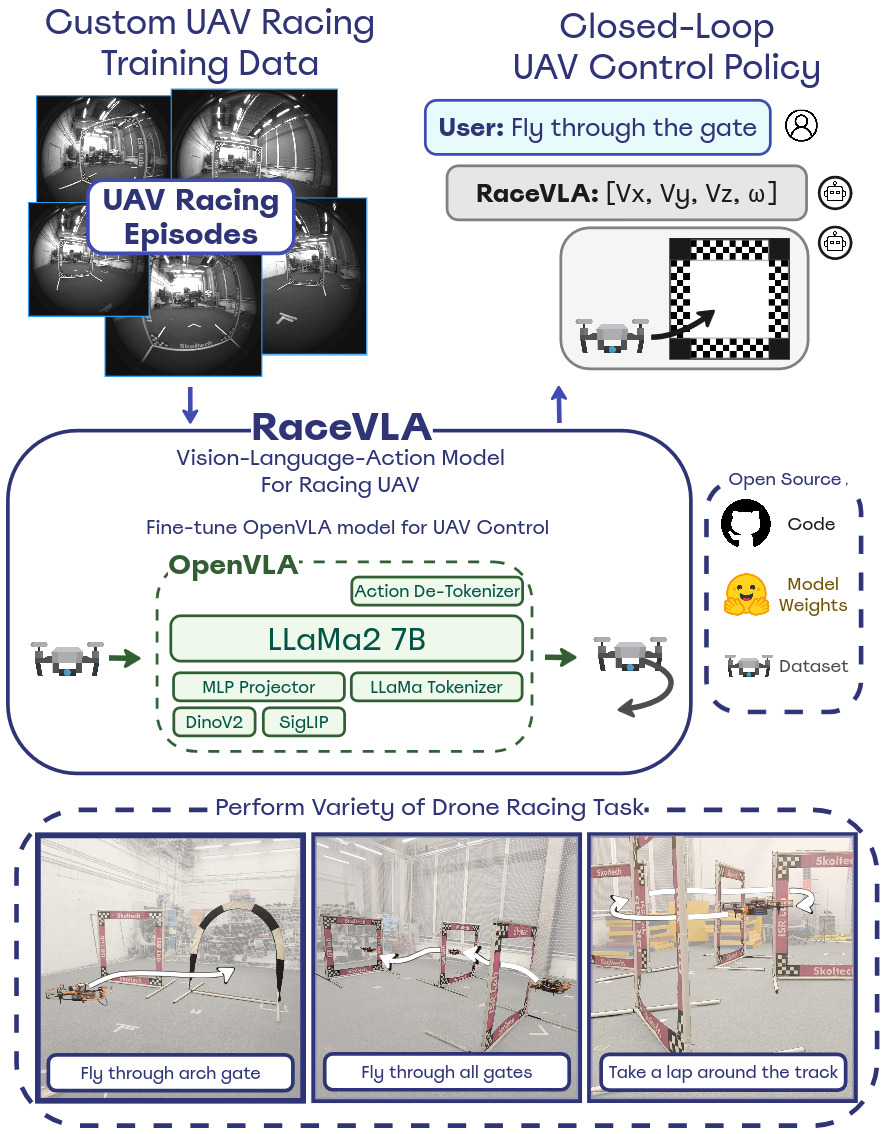}
    \caption{RaceVLA is the first VLA model specifically designed for racing drones. It processes First-Person View (FPV) video streams alongside natural language commands to generate velocity actions (Vx, Vy, Vz) and yaw anglular speed ($\omega$) control signals. This innovative system enables drones to autonomously execute a wide range of flight tasks, including the navigation in novel scenarios in unfamiliar environments. By leveraging a purpose-built training dataset, RaceVLA exhibits robust generalization capabilities.}
    \label{fig:teaser}
\end{figure}

Cognitive robotics is a rapidly evolving field that empowers robots to perform complex tasks defined by natural language in dynamic, real-world environments. This advancement spans a diverse range of robotic platforms, including humanoids, quadrupeds, mobile robots, manipulators, drones, and other types of autonomous systems. While significant progress has been made across these various domains, drones—a vital class of autonomous robots—remain underexplored in certain areas, particularly with respect to the application of Vision-Language-Action (VLA) models.

VLA models represent a powerful paradigm in which control signals are generated directly from multimodal inputs, including visual representations and natural language commands. Unlike traditional approaches, which typically involve generating trajectories, detecting objects, implementing high-speed control, or providing high-level instructions for separate modules to execute, VLA models seamlessly integrate sensory input to produce actionable control signals. This direct mapping not only simplifies the architecture but also allows for unprecedented generalization, enabling robots to tackle tasks that were not part of the training dataset. However, many current AI systems for drones still rely on transformer-based models focused primarily on planning, trajectory generation, or skill selection. These approaches, while effective in certain contexts, tend to limit the adaptability and versatility of drones in dynamic, real-world scenarios.

In this work, we introduce RaceVLA, the first VLA-based system specifically designed for autonomous racing drones. Our model leverages First-Person View (FPV) video streams from onboard cameras and natural language commands to generate a four-component control signal consisting of velocity vector ($V_x$, $V_y$, $V_z$, $\omega$). This novel approach paves the way for the development of universal drones capable of autonomously performing a wide variety of flight tasks, including those previously unseen, in unfamiliar environments. By training the drone on a diverse set of actions, it can generalize across different scenarios, similarly to how VLA models have enabled robotic manipulators to achieve broad versatility in their tasks.

To validate the effectiveness of our system, we utilize racing drone fields as a testing environment. These fields require drones to navigate complex tracks, avoid collisions, and pass through gates. RaceVLA operates in real time, exhibiting human-like decision-making capabilities in controlling racing drones. This breakthrough marks a significant step in applying VLA models to aerial robotics, offering new possibilities for drone autonomy in dynamic and challenging environments.

\section{Related Work}
The field of cognitive robotics has grown rapidly, with significant contributions across various robotic platforms. Quadruped robots such as CognitiveDog~\cite{lykov2024cognitivedoglargemultimodalmodel} and DoggyBot~\cite{wu2024helpfuldoggybotopenworldobject} have enhanced mobility and interaction through sophisticated decision-making processes. Humanoid robots, including Tesla Optimus and Agility Robotics~\cite{Agility_Robotics}, have increasingly incorporated generative AI to improve performance in human-centered environments. Mobile robots, such as LLM-MARS~\cite{lykov2023llmmarslargelanguagemodel}, showcase the versatility of LLM-powered systems in navigation and exploration tasks.

Robotic manipulators have particularly benefited from the application of VLAs. Systems like PaLM-E~\cite{driess2023palmeembodiedmultimodallanguage}, RT-1~\cite{brohan2023rt1roboticstransformerrealworld}, RT-2~\cite{brohan2023rt2visionlanguageactionmodelstransfer}, and RT-X~\cite{embodimentcollaboration2024openxembodimentroboticlearning} are all VLA-based models that enable robots to directly generate control signals from multimodal inputs. These models enhance robots' ability to generalize across tasks and perform actions beyond those presented in their training data. OpenVLA~\cite{kim2024openvlaopensourcevisionlanguageactionmodel} introduced fine-tuning capabilities that allow these models to be adapted for diverse tasks and robotic platforms using consumer-grade hardware, while MiniVLA~\cite{belkhale2024minivla} further reduced model size, making the technology more accessible and useful for inboard inference. The application of transformer-based models to drone systems has also been explored. UAVs Meet LLMs~\cite{tian2025uavsmeetllmsoverviews} provides a comprehensive overview of transformer-based approaches, focusing on their use in drone navigation
~\cite{fan2023aerialvisionanddialognavigation}
~\cite{gao2024aerialvisionandlanguagenavigationsemantictopometric} ~\cite{EmbodiedCityzhang2024} ~\cite{wang2024realisticuavvisionlanguagenavigation}
~\cite{liu2023aerialvlnvisionandlanguagenavigationuavs}, flight control ~\cite{vemprala2023chatgptroboticsdesignprinciples} ~\cite{zhong2023safervisionbasedautonomousplanning}, and mission planning ~\cite{pan2024vlp}, ~\cite{sautenkov2025uavvlavisionlanguageactionlargescale}.

The Aerial Vision Language Navigation has seen significant advancements, particularly within the context of the human-centered behaviour. Initially, Liu et al.\cite{liu2023aerialvlnvisionandlanguagenavigationuavs} introduced the concept of Aerial Visual Language Navigation, alongside the development of the Aerial Visual Language Navigation (VLN) technique and the accompanying AerialVLN dataset. In a related study, Fan et al. \cite{fan2023aerialvisionanddialognavigation} detailed the creation of a simulator and a Visual Language Dialog Navigation (VLDN) system, which enables real-time communication between the drone and its operator during flight operations. Zhang et al. \cite{EmbodiedCityzhang2024} took this innovation further by designing a versatile environment for embodied intelligence in urban settings. Their system allows agents to concurrently perform both VLA and VLN tasks in an interconnected manner. In another notable contribution, Gao et al. \cite{gao2024aerialvisionandlanguagenavigationsemantictopometric} proposed a method where the map is represented as a matrix and fed into a Large Language Model (LLM). They introduced the Semantic Topo Metric Representation (STMR), a technique that facilitates the integration of spatial data into LLMs. Finally, Wang et al. \cite{wang2024realisticuavvisionlanguagenavigation} unveiled the OpenUAV platform, a comprehensive benchmark and simulator offering realistic environments, flight simulations, and extensive algorithmic support for UAV navigation tasks.

Deep reinforcement learning (DRL) \cite{Scaramuzza} has emerged as a transformative approach in champion-level drone racing, pushing the boundaries of autonomous UAV performance. Autonomous drone racing demands exceptional agility, real-time decision-making, and precise trajectory optimization. Additionally, models for drone swarm control, such as SwarmGPT~\cite{jiao2023swarmgptcombininglargelanguage} and FlockGPT~\cite{lykov2024flockgptguidinguavflocking}, have been developed. SwarmGPT generates individual trajectories for each drone in a swarm, while FlockGPT computes Signed Distance Function (SDF) to define the flight direction for each drone based on its position relative to the target surface. While these approaches offer valuable insights, they focus primarily on planning and trajectory generation, which restricts their adaptability and real-time control capabilities. These methods also struggle to generalize to tasks outside their predefined dataset and may fail to quickly adapt to changing environments or respond effectively to unexpected situations.

In contrast, RaceVLA introduces a novel application of VLA to drones, enabling real-time, human-like navigation by processing FPV video inputs and natural language commands. Our work builds upon the foundations of previous VLA models, addressing the limitations of existing approaches by providing greater adaptability, real-time responsiveness, and the ability to generalize to new, unforeseen tasks. This advancement opens the door to more versatile and autonomous drone systems capable of navigating complex and dynamic environments with unprecedented efficiency.

\section{System Architecture of RaceVLA}

\subsection{The VLA Model}

\begin{figure}
    \centering
    \includegraphics[width=1.0\linewidth]{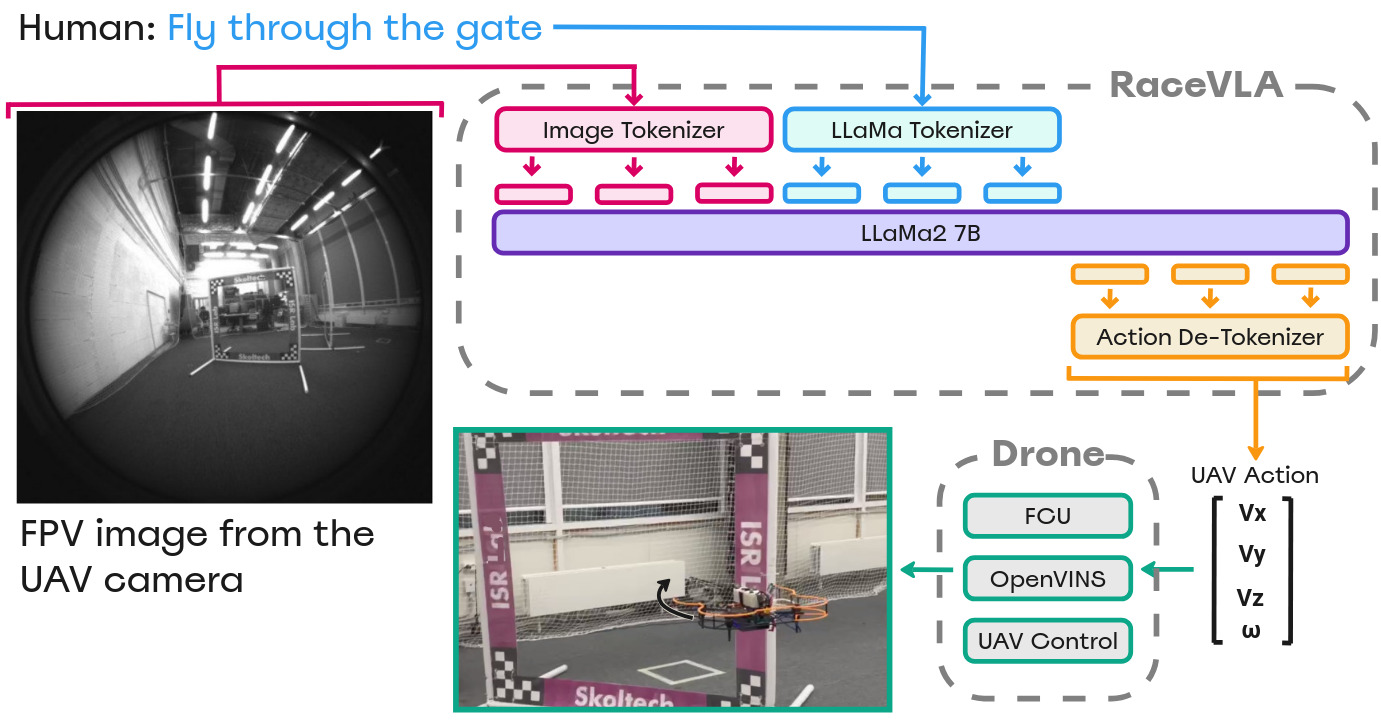}
    \caption{RaceVLA system architecture.}
    \label{fig:sys_arc}
    \vspace{-5mm}
\end{figure}

The core of the RaceVLA system, illustrated in Figure~\ref{fig:sys_arc}, is a VLA model designed for controlling the flight of racing UAVs. Our approach builds upon the OpenVLA model developed by Stanford AI Lab, fine-tuning it on a dataset of racing drone flight episodes. Each sample in the dataset corresponds to a specific task performed by a remotely controlled racing drone, ensuring the model adapts to the high-speed and dynamic nature of drone racing.

The fundamental architecture of the VLA model remains consistent with OpenVLA, maintaining its multimodal input-output structure. The input consists of an FPV frame from the drone’s onboard camera and a natural language instruction. The output is an action vector directly controlling the drone. While OpenVLA was originally trained for robotic manipulation and produced a 7D action vector—three translational deltas, three rotational deltas, and a gripper delta — RaceVLA modifies this to a 4D action vector suitable for drone control. Specifically, our model outputs three linear velocities ($V_x$, $V_y$, $V_z$) and yaw anglular speed ($\omega$). 

Drone navigation under the RaceVLA system follows an iterative process. Given a task description, the drone continuously processes FPV frames and natural language instructions to compute and execute the next flight step. Unlike traditional waypoint-based approaches, RaceVLA does not wait for the drone to reach the specified point before processing the subsequent frame. Instead, the system immediately analyzes the next frame and adjusts the drone's trajectory in real time. This iterative and continuous operation ensures smooth, uninterrupted flight without pauses between steps—a critical improvement for racing drones, where seamless transitions are essential.

To enable real-time processing, we optimized the VLA model by quantizing it to int8 precision. Combined with a high-performance NVIDIA RTX 4090 GPU on the server and efficient API communication, the system achieves an operational frequency of 4 Hz. This balance between computational efficiency and responsiveness ensures RaceVLA can handle the demands of high-speed drone racing, providing smooth and precise flight control.

\subsection{Autonomous Racing Drone}

The proposed system utilizes a custom-built 8-inch racing drone designed for high-speed operation and navigation in dynamic environments. The drone is equipped with a Speedy Bee F405 flight controller running ArduPilot firmware, with control and communication facilitated via MAVROS. A RealSense T265 camera serves as the FPV camera, capturing visual data that is input into the VLA model. The VLA model operates on a stationary PC, receiving real-time frames from the drone via a Flask API, processing them with an inference time of 0.25 s per frame, and sending navigation commands back to the drone. An onboard Intel NUC computer handles the drone's processing tasks, including localization and control. The localization is provided by the OpenVINS visual-inertial odometry system \cite{9196524}, which uses images from the RealSense T265 camera in combination with IMU measurements to accurately track the position of the drone. The entire system operates within the Robot Operating System (ROS) framework, ensuring smooth data flow and integration. The output from the VLA model on the stationary PC is seamlessly incorporated into the drone's onboard navigation and control pipeline, enabling high-performance autonomous flight.

\begin{figure}
    \centering
    \includegraphics[width=\linewidth]{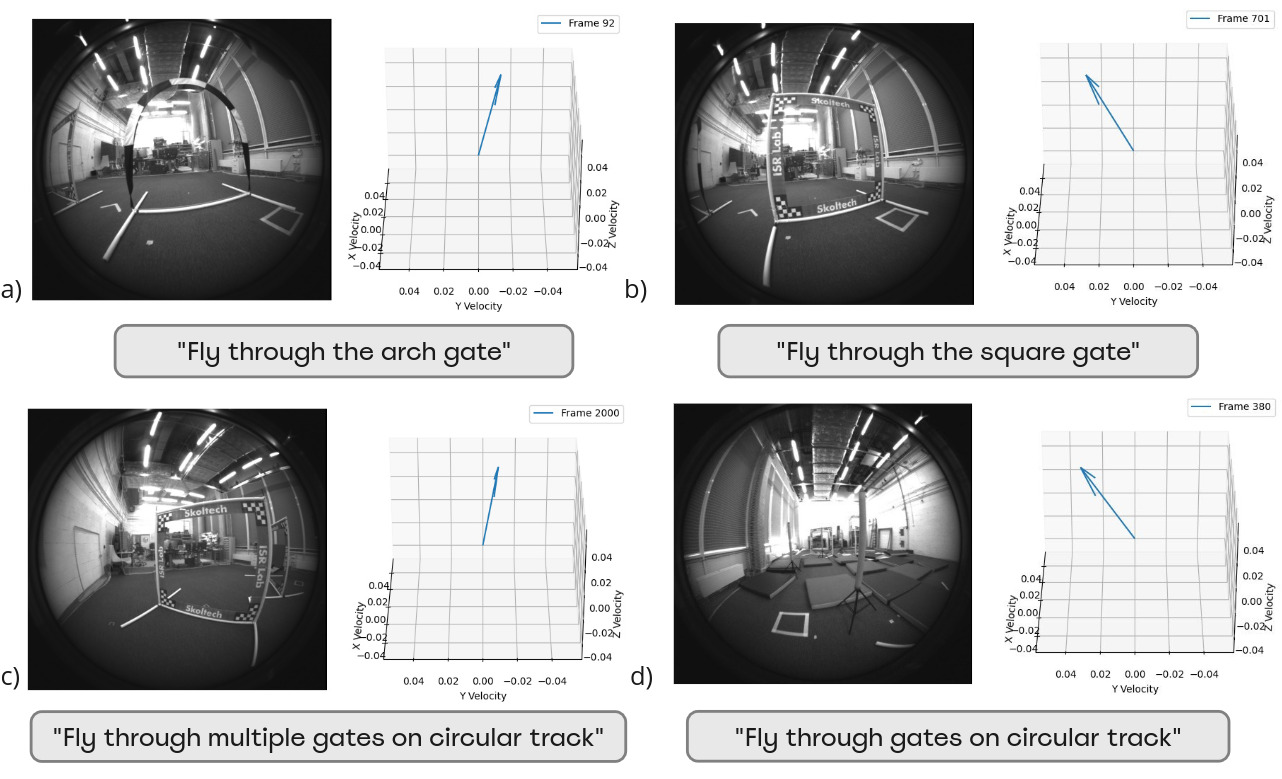}
    \caption{a) Dataset for the task "Fly through the arch gate" featuring trajectories and actions for navigating arch-shaped gates. b) Dataset for the task "Fly through the square gate" highlighting data specific to square-shaped gates. c) "Fly through multiple gates on circular track". d) "Fly through gates on circular track".}
    \label{fig:dataset}
\end{figure}

\section{Dataset Collection and Training Pipeline}
The dataset comprises \textbf{200 episodes} and approximately \textbf{20,000 images}, capturing various racing tasks such as navigating through arch gates, square gates, and circular track formations (see Fig. \ref{fig:dataset}). During data collection, the drone's velocities were recorded using the Vicon motion capture system at a frame rate of \(60\) Hz, while the T265 camera captured images at \(30\) Hz. Each data sample includes positional data, velocity components ($V_x$, $V_y$, $V_z$), yaw angle changes, and synchronized visual frames. Language-based task instructions such as “Fly through the arch gate” and “Fly through multiple gates on circular track” were provided to guide the drone through predefined racing environments.

The dataset was formatted using the Reinforcement Learning Dataset Specification (RLDS), ensuring compatibility with OpenVLA for efficient structuring of actions, images, and instructions. The collected data was used to fine-tune the OpenVLA-7b model, a \(7\)-billion parameter vision-language-action model. A parameter-efficient Low-Rank Adaptation (LoRA) technique with rank-\(32\) adapters was employed to optimize memory usage while training a minimal set of model weights.

The training pipeline was configured with a batch size of \(16\), a learning rate of \(5 \times 10^{-4}\), and \(7000\) gradient steps, with image augmentation disabled to preserve data authenticity. Training was conducted on a single NVIDIA A100 GPU to handle the computational load of processing the multimodal dataset. Throughout the training process, key performance metrics such as training loss, action prediction accuracy, and L1 loss were monitored. Smoothed metrics were applied to ensure stable and reliable evaluation. The training process was managed using Weights \& Biases (W\&B), enabling real-time tracking of model performance and periodic checkpointing for evaluation. The fine-tuned model aims to achieve real-time action prediction by leveraging visual input and language commands for high-speed drone navigation with minimal computational overhead.
\section{Experiments}

\subsection{VLA Evaluation on Drone Racing}

The RaceVLA was evaluated on drone racing tasks, requiring navigation and flight through gates on a racing track. The experiments aimed to assess the model's performance in various flight scenarios.

\begin{figure}
    \centering
    \includegraphics[width=0.9\linewidth]{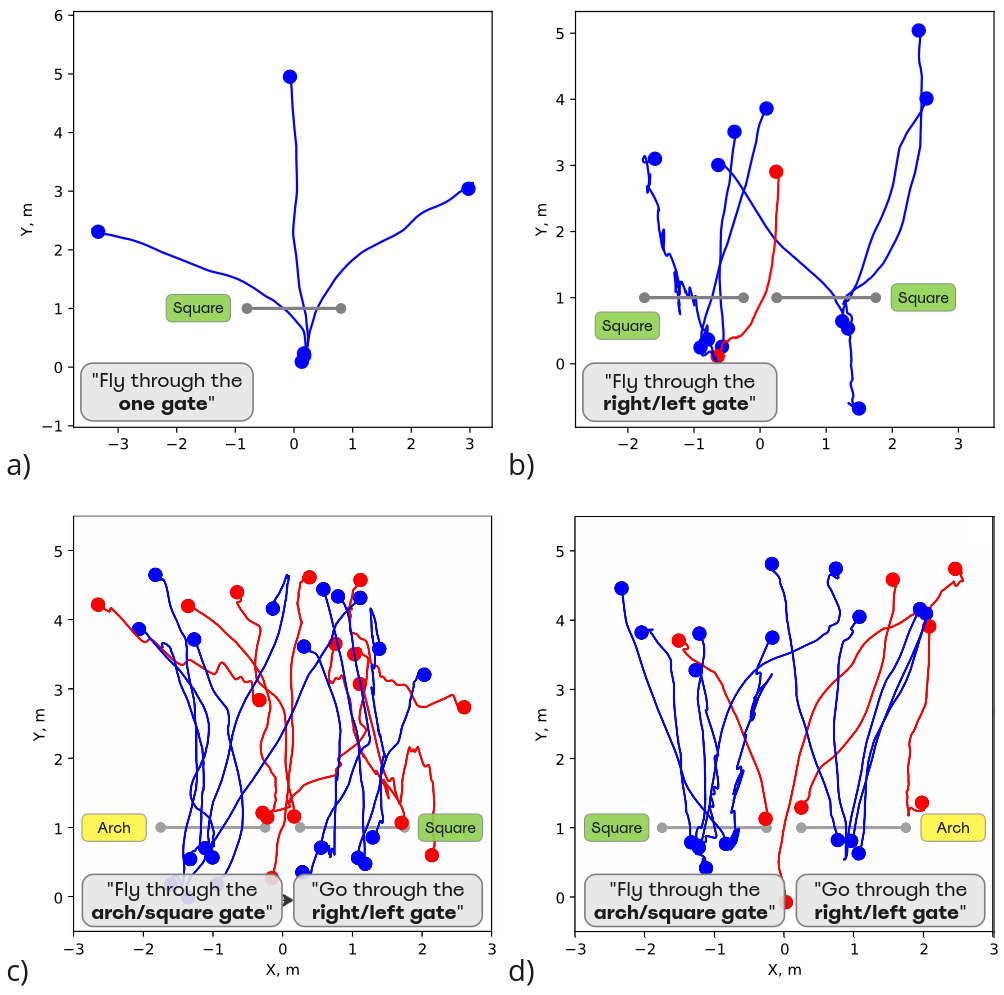}
    \caption{Evaluation of the RaceVLA system for autonomous drone navigation through racing gates, starting from different initial positions (seen, unseen positions). (a) Performance of the model starting the drone from different initial positions with the task "Fly through one gate." (b) Model evaluation for tasks "Fly through the Right gate" and "Fly through the Left gate." (c) Evaluation for sequential tasks "Fly through the Arch gate," "Fly through the Square gate," "Fly through the Right gate," and "Fly through the Left gate." (d) Evaluation in a scenario where the Arch gate is positioned on the right side of the flight zone, and the Square gate is on the left side of the flight zone.}
    \label{fig:4_plots}
\end{figure}

The first test evaluated the completion of tasks illustrated in Figures \ref{fig:4_plots} a, b, c, and d, which included navigating through a single gate, selecting specific gates based on commands, and completing sequences of multiple gates under varying configurations. The flight parameters for this experiment revealed an average speed of 0.51 m/s, a maximum speed of 0.80 m/s, and a standard deviation of speed of 0.17 m/s. The average yaw angular velocity was 0.243 rad/s. The average duration of the tasks is 26.92 s.

\begin{figure}
    \centering
    \includegraphics[width=1.0\linewidth]{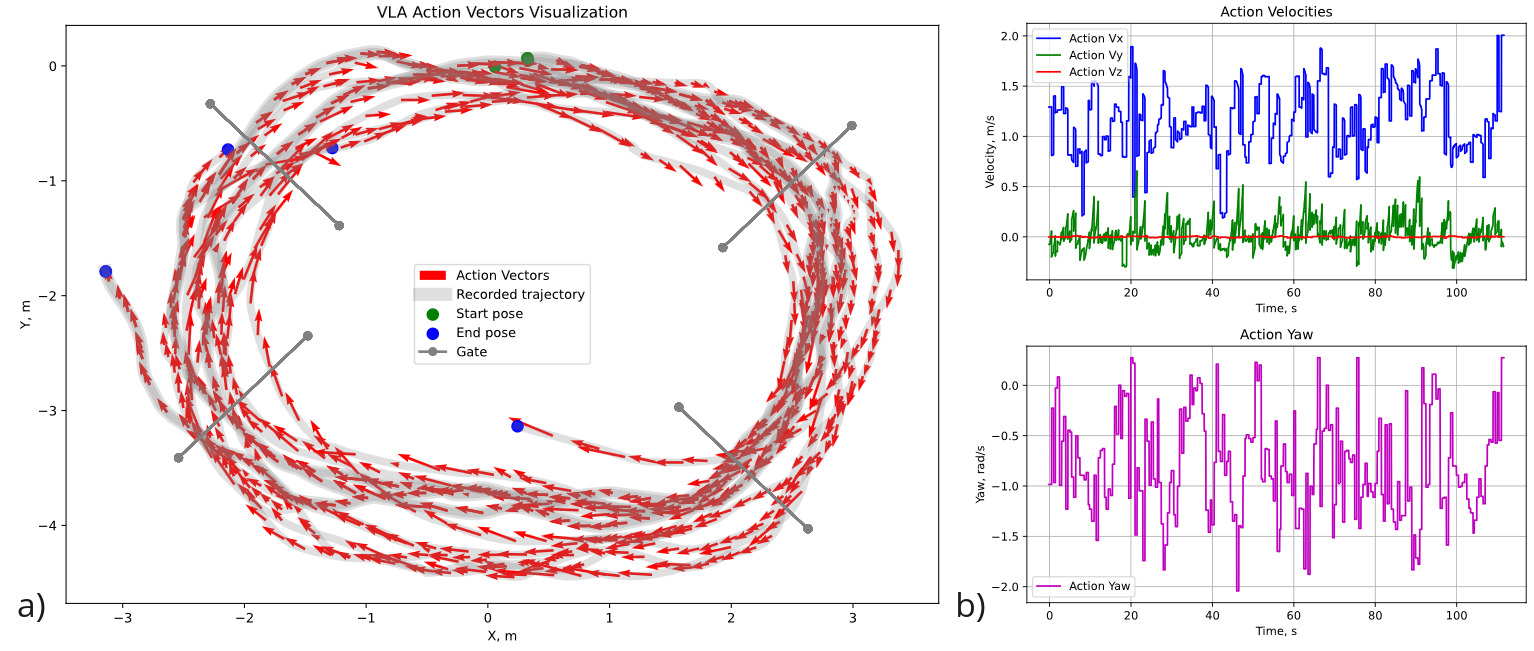}
    \caption{a) Plots of recorded trajectories in the circular track task. The recorded trajectory (gray), actions generated by the VLA model (red arrows), and drone racing gates (gray lines) are shown. b) The right plot visualizes the velocity action vector and yaw rotation of the drone for 3 laps.}
    \label{fig:traj_lap}
    \vspace{-4mm}
\end{figure}

The second test consisted of a series of flights along a predefined racing track. Figure \ref{fig:traj_lap} presents the plots of recorded one of the trajectory during the "Flying through multiple gates on circular track" task. Across all three trajectories, the average mean velocity is approximately 1.04 m/s, with a maximum velocity of 2.02 m/s. The standard deviation of velocity is 0.27 m/s, indicating some variation in speed across the flights. The average mean yaw angular velocity is 0.40 rad/s. The angular velocities show moderate variation, as indicated by the standard deviation of 0.12 rad/s. 

\subsection{Evaluation on VLA Generalization}

To evaluate RaceVLA’s generalization capabilities, we followed the experimental framework established in OpenVLA, adapting it to the unique challenges of drone racing. The evaluation was designed to assess generalization across four distinct axes: visual, motion, physical, and semantic. These axes were tested through a comprehensive set of tasks, involving 200 experiments conducted with RaceVLA. 

The evaluation was structured as follows:
\begin{itemize}
    \item \textbf{Visual Generalization:} Tasks were designed to test how well RaceVLA generalizes to new environments with unseen backgrounds, distractor objects, and variations in the appearance of target items.
    \item \textbf{Motion Generalization:} New racing tracks were created to evaluate RaceVLA's ability to handle unseen object positions and orientations during dynamic motion.
    \item \textbf{Physical Generalization:} Tasks included modifications to the shape and size of gates, testing RaceVLA’s ability to adapt to changes in the physical environment.
    \item \textbf{Semantic Generalization:} Unseen drone-specific tasks were set to evaluate the model's ability to comprehend and act upon unfamiliar instructions and target objects.
\end{itemize}

Unlike the static manipulators in OpenVLA, the racing drones in RaceVLA operate with dynamic cameras, which added complexity to the environment. The drone’s camera continuously captures new input, with changing object positions and backgrounds, providing diverse visual data for training and evaluation. This distinction from static camera systems required an adaptation of the OpenVLA experimental framework but maintained the core principles for a fair comparison.

\textbf{Results and Analysis:}

\begin{figure}[t!]
    \centering
    \includegraphics[width=1.0\linewidth]{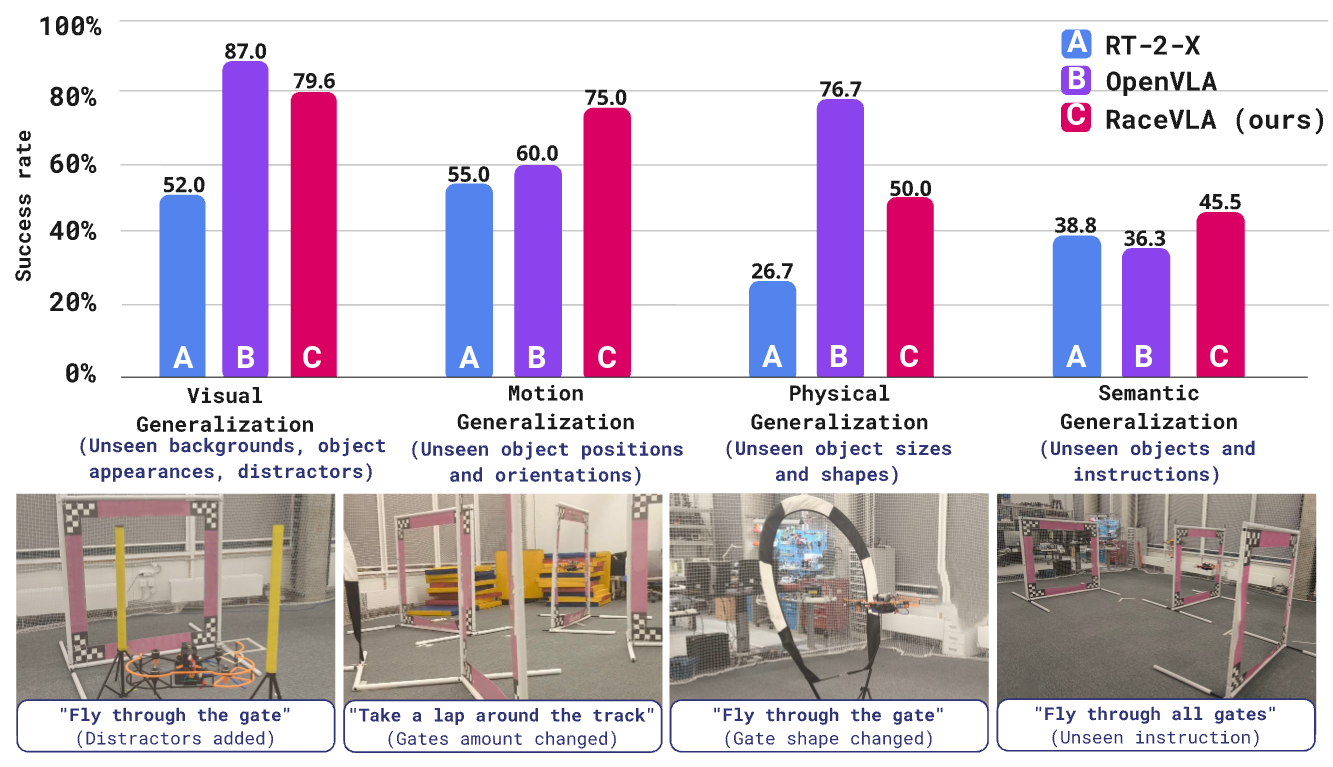}
    \caption{The figure presents RaceVLA’s generalization evaluation, with a comparison to benchmarks shown at the top and examples of experimental scenarios provided below for each of the four generalization axes: visual, motion, physical, and semantic.}
    \label{fig:generalixation}
    \vspace{-4mm}
\end{figure}

The generalization performance of RaceVLA across four axes — visual, motion, physical, and semantic — shown in Figure~\ref{fig:generalixation}, includes both the evaluation results and task examples for different generalization axes. RaceVLA demonstrated superior performance compared to OpenVLA in motion (75.0 vs. 60.0) and semantic (45.5 vs. 36.3) generalization. The dynamic nature of the drone's camera played a pivotal role in this improvement, as it provided diverse visual input within individual dataset samples. Furthermore, the absence of object manipulation tasks in RaceVLA simplified the motion dynamics, enabling the model to excel in scenarios requiring simpler actions, such as navigating around new objects rather than grasping them. This simplification also contributed to enhanced semantic generalization by eliminating the need for complex physical interactions with objects.

Despite these advantages, RaceVLA exhibited slightly reduced performance compared to OpenVLA in visual (79.6 vs. 87.0) and, more noticeably, physical (50.0 vs. 76.7) generalization. The drone's navigation in dynamic, visually complex environments likely contributed to this disparity. Unlike static manipulators, drones must avoid distractors by physically maneuvering around them, which increases the complexity of maintaining visual focus. Additionally, in physical generalization tasks, where gate shapes and sizes vary significantly, the drone’s spatial perspective introduced challenges. Objects in the drone’s field of view often appeared smaller or more distorted due to distance or motion, potentially disrupting its spatial understanding and impacting its ability to adapt to shape variations effectively.

Notwithstanding these challenges, RaceVLA demonstrated a marked advantage over RT-2 by Google DeepMind across all generalization axes, including visual (79.6 vs. 52.0), motion (75.0 vs. 55.0), physical (50.0 vs. 26.7), and semantic (45.5 vs. 38.8). These results underscore RaceVLA's robustness in adapting to dynamic and complex environments, outperforming state-of-the-art models in domains requiring real-time adjustments to both physical and visual stimuli.

Overall, RaceVLA showcases remarkable potential for applications in highly dynamic settings, such as drone racing, where interaction with unpredictable and rapidly changing environments is critical. While challenges remain in optimizing physical and visual generalization under these conditions, the model's ability to excel in motion and semantic tasks highlights its adaptability and promise for advancing autonomous systems in real-world scenarios.

\section{Conclusion}
RaceVLA represents a novel approach to autonomous drone racing navigation, leveraging the VLA model to replicate human-like decision-making in dynamic environments. RaceVLA outperforms RT-2 across all generalization aspects and even surpasses the state-of-the-art OpenVLA in specific axes, such as motion (\textbf{75.0} vs. \textbf{60.0}) and semantic generalization (\textbf{45.5} vs. \textbf{36.3}). The dynamic nature of the drone's camera trains the model to be independent of camera positioning, driving significant improvements in motion generalization. While RaceVLA demonstrates impressive capabilities, increasing the drone's speed will require reducing the model's inference time to ensure timely navigation updates. Furthermore, the task-specific focus of racing drones, which emphasizes navigation around objects rather than complex manipulation, contributes to enhanced semantic and motion generalization. As a result, RaceVLA demonstrates significant potential for high-performance and highly environment and task adaptive navigation and sets a new benchmark for VLA drone performance. Future work will focus on improvement of visual and physical generalization in more unpredictable and diverse racing scenarios. Furthermore, expanding the data set and refining the model’s ability to handle a wider range of environmental conditions could further improve its robustness and adaptability. The findings suggest that RaceVLA could play a pivotal role in advancing the field of autonomous drone racing, with future iterations offering even greater precision and efficiency in competitive contexts.

\bibliographystyle{IEEEtran}
\bibliography{ref} 
\balance
\end{document}